\documentclass[letterpaper]{article} 
\usepackage{aaai23}  
\usepackage{times}  
\usepackage{helvet}  
\usepackage{courier}  
\usepackage[hyphens]{url}  
\usepackage{graphicx} 
\usepackage{natbib}  
\usepackage{caption} 
\usepackage{algorithm}
\usepackage{algorithmic}
\usepackage{footmisc}

\usepackage{algorithm}
\usepackage{algorithmic}
\usepackage{tabularx}
\usepackage{xparse}
\usepackage{amsmath}
\usepackage{amsthm}
\usepackage{amssymb}
\usepackage{amsfonts}
\usepackage{xspace}
\usepackage{relsize}
\usepackage{graphicx}
\usepackage{multirow}
\usepackage{url}
\usepackage{subfigure}
\usepackage{comment}
\usepackage{xcolor}
\newcommand{\indep}{\perp \!\!\! \perp}

\usepackage{newfloat}
\usepackage{listings}
\DeclareCaptionStyle{ruled}{labelfont=normalfont,labelsep=colon,strut=off} 
\floatstyle{ruled}
\newfloat{listing}{tb}{lst}{}
\floatname{listing}{Listing}
\title{Deep Learning on a Healthy Data Diet: Finding Important Examples for Fairness}
\author{
     Abdelrahman Zayed \textsuperscript{\rm 1,2} \footnote{Work done during an internship at Microsoft Research. } ,
     Prasanna Parthasarathi \textsuperscript{\rm 1,3},
     Gonçalo Mordido \textsuperscript{\rm 1,2},
     Hamid Palangi \textsuperscript{\rm 4},
     Samira Shabanian \textsuperscript{\rm 4} \thanks{Equal advising.} ,
     Sarath Chandar \textsuperscript{\rm 1,2,5} \footnotemark[2] 
}

\affiliations{
    \textsuperscript{\rm 1}Mila - Quebec AI Institute,
    \textsuperscript{\rm 2}Polytechnique Montreal,
    \textsuperscript{\rm 3}McGill University,
    \textsuperscript{\rm 4}Microsoft Research,
    \textsuperscript{\rm 5}Canada CIFAR AI Chair\\

    \{zayedabd,parthapr,goncalo-filipe.torcato-mordido,sarath.chandar\}@mila.quebec
    \{hpalangi,samira.shabanian\}@microsoft.com
}

\usepackage{bibentry}

\begin{document}

\maketitle

\begin{abstract}
Data-driven predictive solutions predominant in commercial applications tend to suffer from biases and stereotypes, which raises equity concerns. Prediction models may discover, use, or amplify spurious correlations based on gender or other protected personal characteristics, thus discriminating against marginalized groups. Mitigating gender bias has become an important research focus in natural language processing (NLP) and is an area where annotated corpora are available.  \textcolor{black}{Data augmentation reduces gender bias by adding counterfactual examples to the training dataset. In this work, we show that some of the examples in the augmented dataset can be not important or even harmful to fairness. We hence propose a general method for pruning both the factual and counterfactual examples to maximize the model’s fairness as measured by the demographic parity, equality of opportunity, and equality of odds. The fairness achieved by our method surpasses that of data augmentation on three text classification datasets, using no more than half of the examples in the augmented dataset. Our experiments are conducted using models of varying sizes and pre-training settings. \textit{WARNING: This work uses language that is offensive in nature.}}
\end{abstract}

\section{Introduction}

  Although pre-trained language models \cite{vaswani2017attention,devlin2018bert,radford2018improving,radford2019language,brown2020language,raffel2019exploring, yang2019xlnet} have been tested on a variety of language understanding and generation benchmarks \cite{wang2018glue, wang2019superglue, rajpurkar2016squad, budzianowski2018multiwoz, zhang2018personalizing,  mark-evaluate, sauder-etal-2020-best}, the fairness of these models with respect to marginalized communities has recently come under scrutiny \cite{dixon2018measuring, zhang2020demographics, garg2019counterfactual}. In terms of unintended gender bias in prediction, the works by \citet{nadeem2020stereoset} and \citet{meade2021empirical} show that pre-trained language models, such as BERT \cite{devlin2018bert}, RoBERTa \cite{liu2019roberta}, GPT-2 \cite{radford2019language}, and XLNet \cite{yang2019xlnet} behave differently based on the presence or absence of gender cues in the input text, while the prediction should have remained agnostic. 

Similar results have been shown for non-pretrained recurrent networks \cite{maudslay2019s, de2019bias, lu2020gender} such as long short-term memory models (LSTMs) \cite{hochreiter1997long} and gated recurrent units (GRUs) \cite{chung2014empirical}. \citet{kiritchenko2018examining} also highlight that support vector machines (SVMs) \cite{hearst1998support} applied for sentiment analysis predict \emph{anger} with a higher probability for input texts having references to female gender (\textit{e.g.} \emph{she}, \emph{her}, \emph{woman}, \emph{lady}) over texts with male references (\textit{e.g.} \emph{he}, \emph{him}, \emph{himself}, \emph{man}), with the same context. A biased model should not replace humans in certain tasks (\textit{e.g.} resume filtering or loan eligibility prediction), regardless of how accurate it is if it achieves high accuracy on a test set that is not representative of the population. For example, if most test examples refer to men, this could hide the model’s poor performance on examples that reference women. Hence, relying solely on metrics such as accuracy and $F1$ might be misleading.

To mitigate gender bias in language models, several techniques have been proposed. These methods can be broadly classified into three main categories: data-based methods \cite{lu2020gender,maudslay2019s,de2019bias}; regularization-based methods \cite{gupta2021controllable,garg2019counterfactual}; and adversarial-based methods  \cite{song2019learning,madras2018learning,jaiswal2020invariant}. Data-based methods, which are our focus in this work, change the training data to balance the bias through targeted data augmentation \cite{lu2020gender, maudslay2019s}, constructing counterfactual examples \cite{garg2019counterfactual}, or removing protected attributes from the input \cite{de2019bias} to disallow models from learning any correlation between labels and gender words. Regularization-based methods add an auxiliary loss term to the objective function that reduces the amount of bias in the model. Adversarial-based methods use adversarial learning for bias mitigation. 


Although data-based methods are effective in theory, they are also demanding in different ways. First, some methods such as counterfactual data augmentation (CDA) \cite{lu2020gender} add counterfactual examples to the training data, which substantially increases the training time. Second, data balancing methods \cite{dixon2018measuring} manually collect more examples for the under-represented groups, which requires human intervention. Third,
the performance may degrade on the main downstream task  \cite{zhang2020demographics, meade2021empirical}.
\textcolor{black}{In this work, we propose the gender equity (\textit{GE}) score, which ranks the counterfactual examples based on their contribution to the overall fairness of the model. Intuitively, this score makes data augmentation methods more efficient by only using the examples that have the largest contribution to fairness. The pruning nature of our approach also helps exclude the harmful examples that degrade the overall fairness of the model by enforcing stereotypical associations about different genders. Our goal is to find the best trade-off between fairness and performance.} Our contributions are summarized as follows: 
\begin{enumerate}

\item \textcolor{black}{We propose a way to filter the examples that enforce undesired gender stereotypes from the training data,  thus improving the overall fairness of the model compared to conventional data augmentation.}

\item \textcolor{black}{We reduce the redundancy in counterfactual data augmentation by pruning the counterfactual examples that are not important for fairness (\textit{i.e.} the ones that do not contain gender words).}

\item We study the effect of gender bias mitigation on downstream task performance and find that our method only shows a degradation of no more than $3\%$ on the AUC over three popular language understanding tasks that are concerned with biases and stereotypes, compared to the original (biased) BERT and RoBERTa models.

\end{enumerate}

\section{Related Work}\label{sec:related_work}

\paragraph{Gender bias}
Gender bias is defined as the tendency of the system (which is the machine learning model in our case) to change its prediction based on the gender of the person referred to in the sentence \cite{friedman1996bias}. Existing works that study gender bias can be categorized into structural
\cite{adi2016fine, hupkes2018visualisation, conneau2018you, tenney2019bert, belinkov2019analysis} or behavioural \cite{sennrich2016grammatical,isabelle2017challenge,naik2018stress} approaches. Our work follows a behavioral approach for bias quantification. 
Structural methods measure gender bias by focusing on the embeddings that the model assigns to sentences or words, regardless of the model's prediction \cite{tenney2019bert,belinkov2019analysis}. One of the earliest structural methods to quantify bias is known as the {\bf W}ord {\bf E}mbedding {\bf A}ssociation {\bf T}est (WEAT) \cite{brunet2019understanding}, which measures the similarity between the word embeddings of some target words (such as ``men'' and “women”) and some attribute words (such as “nice” or “strong”). According to this metric, a model is considered biased if the word embeddings for masculine words such as “man” and “boy” have a higher cosine similarity with attribute words such as “strong” and “powerful" than feminine words such as “woman” and “girl”. The work by \citet{may2019measuring} extended this concept to sentence embeddings by introducing a {\bf S}entence {\bf E}mbedding {\bf A}ssociation {\bf T}est (SEAT).

Behavioral methods, on the other hand, measure the model's bias based on its predictions on synthetic datasets designed specifically for gender bias assessment. The work by \citet{dixon2018measuring} measured bias using the {\bf F}alse {\bf P}ositive {\bf E}quality {\bf D}ifference (FPED) and {\bf F}alse {\bf N}egative {\bf E}quality {\bf D}ifference (FNED). In other words, they measure bias as the inconsistency in false positive/negative rates across different genders. For example, if a model has a higher false positive/negative rate when the input examples refer to women, then the model is biased.

\paragraph{Data-based bias mitigation methods}
One of the early works in data-based bias mitigation methods is the work by \citet{dixon2018measuring}, which proposed an expensive, yet effective, data augmentation technique through manual labeling to account for identity balancing issues. \citet{lu2020gender} proposed counterfactual data augmentation (CDA) which requires constructing a counterfactual example for every example in the dataset, thus doubling the size of the training data. \textcolor{black}{\citet{maudslay2019s} proposed counterfactual data substitution (CDS) where the model replaces each example with its corresponding counterfactual example based on a fair coin toss, which keeps the training data size constant}. The authors proposed also to replace names, so they refer to another gender. \citet{de2019bias} proposed not exposing the model to gender tokens by simply removing them from the dataset.

Another similar approach is called counterfactual logit pairing \citep{garg2019counterfactual}, which involves creating a counterfactual sentence for every input sentence, such that the model is encouraged to give the same predictions for both. For any input sentence, the corresponding counterfactual sentence carries the same meaning while referring to an alternative identity group. For example, a sentence like “he is a gay man” may become “he is a straight man”. The intuition here is to teach the model not to base its decision on identity characteristics such as sexual orientation, nationality, and gender. Although the authors empirically show the effectiveness of this method, it is limited by the automatic replacement of a group of $50$ words representing different identity groups \citep{dixon2018measuring}. Among those $50$ words, only the words “male” and “female” are related to gender, which restricts the method's applicability to sentences containing these exact words. For example, gender pronouns, which are strong indicators of gender, are ignored.



\paragraph{Performance-based data pruning methods} \citet{toneva2018empirical} propose ranking the examples of a given training dataset based on the number of times they are forgotten during training, denoted as the \textit{forget score}. Forgetting happens when the example is classified incorrectly after it had been correctly classified in a previous epoch. Intuitively, the examples that are forgotten more often during training are more important and should be kept, while those that are never forgotten could be removed from the dataset without affecting the model's performance. On the other hand, \citet{paul2021deep} propose to rank the training examples based on the $\ell_2$-norm of the difference between the prediction of the model and the ground truth, denoted as the \textit{EL2N} score. The larger the norm of the error, the more important the example is. The same authors also propose to rank the training examples based on the mean of the $\ell_2$-norm of the gradient of the loss function with respect to the weights, denoted as the \textit{GraNd} score. Both scores have a strong correlation and are calculated early in training. 

\section{Background}

We define a text classification task 
on a dataset $D$ of $N$ examples, such that $D=\{\mathrm{s}_0,\mathrm{s}_1,\ldots, \mathrm{s}_{N-1}\}$ where $\mathrm{s}_i$ is the $i^{th}$ example. Each sentence $\mathrm{s}_i$ is a composition of $m$ tokens, where $m$ is the maximum number of tokens in any sentence, such that $\mathrm{s}_i = \left\{w^0_i,w^1_i,\ldots, w^{m-1}_i\right\}$, \textcolor{black}{where $w$ represents the token}. The objective of the task is to learn a classifier, parameterized by $\theta$, to output a label $y_i$ $\in$ $\{0,1\}$, where  $y_i$ is $1$ when $\mathrm{s}_i$ is a toxic/sexist sentence, and $0$ otherwise. The optimal set of parameters $\theta^*$ for the model is found using maximum likelihood, as follows: 
\begin{equation}
    \theta^* = \underset{\theta} {\mathrm{argmax}} \prod_{i=0}^{N-1} \mathcal{L_{\theta}}\left(y_i \mid \phi\left(\mathrm{s}_i\right)\right) 
\end{equation}
where $\phi\left(\mathrm{s}_i\right)$ yields the sentence embedding $\mathrm{x}_i$ from the sentence $\mathrm{s}_i$, such that  $\phi: \mathrm{s}_i \rightarrow \mathrm{x}_i \in \mathbb{R}^{K \times m}$, $K$ is the dimension of the word embedding, and $\mathcal{L_{\theta}}$ is the likelihood of the  predictor function parameterized by $\theta$.


To understand the bias in the classifier, we disentangle the intent of a sentence $\mathrm{s}_i$ from the gender identity denoted by $z_i$. For example, if $\mathrm{s}_i|do(z=z_{i})$ is ``he explained the situation to her'' then $\mathrm{s}_i|do(z=\neg z_{i})$ would be ``she explained the situation to him'', where $\mathrm{s}_i|do(z=\neg z_{i})$ represents the same sentence $s_{i}$  had the gender words been flipped \cite{pearl1995}. This is referred to as the counterfactual sentence. Based on the previously mentioned notations, a model, parameterized by $\theta$, is considered biased if its output $y$ depends on $z$, \textit{i.e.}, $y \not\indep z$.  


\section{Deep Learning on a Healthy Data Diet}

Building on the data diet approach introduced by \citet{paul2021deep}, we propose a ``healthier'' version of the data diet that optimizes both fairness and performance using a reduced set of examples. The proposed \textit{GE} score is inspired by the error norm score (\textit{EL2N}) \cite{paul2021deep}, which determines the importance of an example for performance by how far its prediction is from the ground truth. Similarly, we determine the importance of an example for fairness by
how far its prediction is from the prediction when the gender words in the input sentence are flipped. We do so by estimating the norm of the difference between the prediction for factual examples and the corresponding counterfactual examples. Our score is defined as:
\begin{equation} 
\textit{GE}(\mathrm{s}_i) = || f_{\theta}\left(\mathrm{s}_i|do(z=z_{i})\right) -  f_{\theta}\left(\mathrm{s}_i|do(z=\neg z_{i})\right) ||_{2}
\label{eq:our_fairness_score}
\end{equation}
where $f_{\theta}\left(\mathrm{s}_i\right)$ represents the logit outputs of the model for the $i$-th input sentence $\mathrm{s}_i$ and $\|\cdot\|_2$ is the $\ell_2$-norm. The $do(\cdot)$ indicates whether the logits are obtained using the actual gender indication ($z = z_i$) or its counterfactual version ($z= \neg z_i$). Note that examples that do not contain gender words will get a score of zero since we assume they do not contribute to the model fairness with respect to gender. The intuition behind the \textit{GE} score is that if the model's prediction changes drastically upon changing the gender words in a sentence, then this reveals that the sentence contains gender words that the model correlates with the output label. In this case, we add the counterfactual example with the same ground truth label as the factual example to help mitigate this undesirable effect by teaching the model that the output should remain the same independently of the gender words.


 It is important to mention that we flip names (for instance, \textit{John} to \textit{Alice}), gender pronouns, as well as gendered words such as \textit{king} and \textit{queen}. Based on this definition, our \textit{GE} score for any sentence $s_i$ is the same as the score for its counterfactual. Although gender is not binary \cite{manzini2019black, dinan2020multi}, we assume it to be binary in this work for simplicity. We intend to extend our method to non-binary gender in future work.


\subsection{Finding important counterfactual examples for fairness}

\label{sec:finding_minimum_examples}
The \textit{GE} score is used to measure the importance of every counterfactual example in the dataset for fairness. With this information, we propose to train with only the most important examples for fairness to perform bias mitigation. Similar to the \textit{EL2N} score, the \textit{GE} score is computed during the early stages of training and averaged over multiple initializations. The number of epochs used to compute our score is a small fraction of the total number of epochs needed for convergence. It is important to note that \textit{GE} score is both dataset and model-dependent. We study the effect of changing the dataset, model architecture and initialization, as well as the number of the early stage epochs, after which the score is computed in section A.1 of the technical appendix.

\subsection{Combining the factual and counterfactual examples}
\label{sec:combining_our_score_with_other_scores}
\begin{figure}[h]
     \centering
    \centering
    \includegraphics[width=\linewidth]{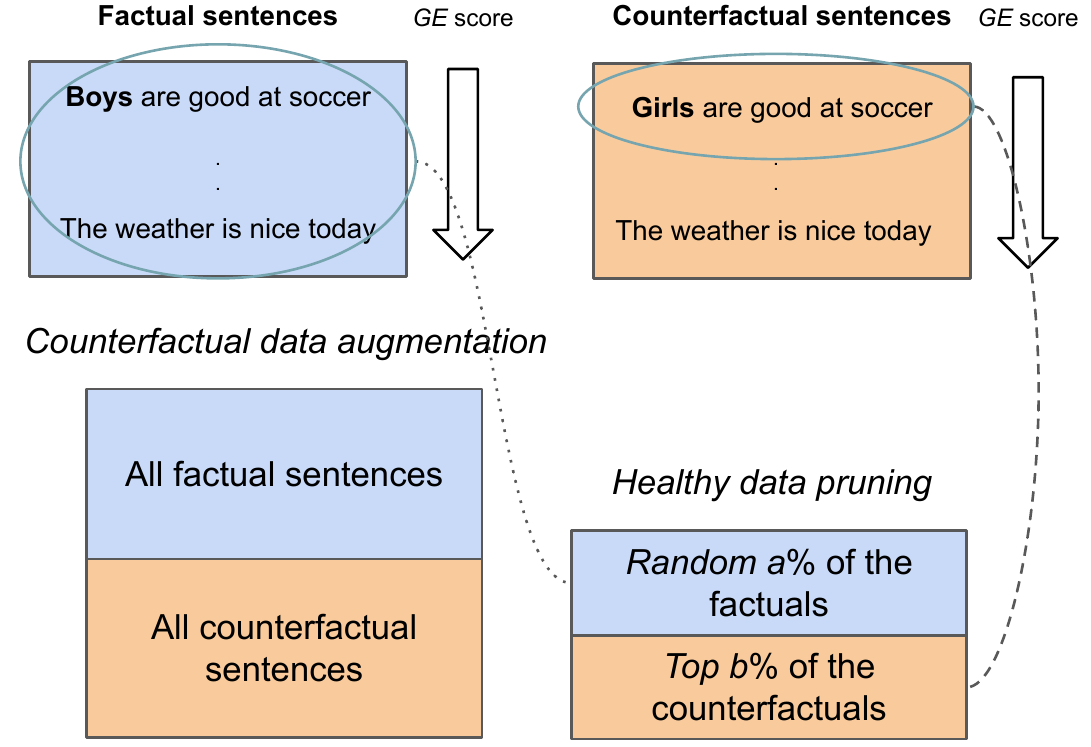}
    \caption{\textcolor{black}{A comparison between our healthy pruning method and conventional CDA, which uses all the counterfactual and factual examples. Healthy data pruning reduces the number of examples needed to promote fairness. The counterfactual examples with high $GE$ scores mitigate the stereotypical correlations that the model might have learned during pre-training, so we prioritize adding them.}
    }
    \label{fig:healthy_data_diet}
\end{figure}
\textcolor{black}{CDA includes both the factual as well as the counterfactual examples in the training dataset, while CDS proposes replacing the factual version of every example with its counterfactual one with a probability $0.5$. In this work, we provide a recipe for a novel way of combining the factual and counterfactual examples such that we outperform the fairness obtained by CDA and CDS, as measured by the demographic parity (DP), equality of opportunity for $y=1$ (EqOpp1), and equality of odds (EqOdd).}

\textcolor{black}{We choose $a\%$ and $b\%$ from the factual and counterfactual examples, respectively, and prune the rest. The $a\%$ from the factual examples are chosen randomly, while the $b\%$ chosen from the counterfactual examples are the ones with the highest $GE$ score, as shown in Fig. \ref{fig:healthy_data_diet}. The intuition is that the counterfactual examples with high $GE$ scores are mostly the ones that mitigate the stereotypical biases that the model might have learned during pre-training \cite{vig2020investigating, nadeem2020stereoset}, so they are useful for fairness.  We repeat this for different combinations of $a\%$ and $b\%$ to find the one that achieves the highest fairness.}

\section{Experiments}
In this section, we describe the tasks, datasets, baselines, and evaluation metrics used in our experiments. Overall, we showcase how our proposed \textit{GE} score reflects the importance of the counterfactual examples on multiple models and datasets. Moreover, we study how \textit{GE} score may be leveraged to achieve a good trade-off between performance and fairness.


\subsection{Datasets}\label{datasets}
We consider two different binary text classification tasks --- sexism and toxicity detection---. The tasks are to train a model to appropriately distinguish texts that are sexist/toxic from the ones that are not. Following \citet{dixon2018measuring}, a toxic comment is defined as a comment that leads a person to leave a discussion. In all previously mentioned tasks, the model should base its predictions on the meaning of the sentence, rather than the gender of the person it refers to. We use the following three datasets:
\begin{enumerate}
\item Twitter dataset \cite{waseem2016hateful}:
This dataset is composed of approximately $16$K tweets, that are classified as racist, sexist, or neither racist nor sexist. We binarized the labels to only sexist and not sexist, by merging the racist tweets with the non-sexist tweets.
\item Wikipedia dataset \cite{dixon2018measuring}:
This dataset is composed of approximately $160$K comments labeled as toxic or not toxic, which are extracted from Wikipedia Talk Pages \footnote{\url{https://figshare.com/articles/dataset/Wikipedia_Talk_Labels_Toxicity/4563973}}.
\item Jigsaw dataset:
This dataset is composed of approximately $1.8$M examples taken from a Kaggle competition\footnote{\url{https://www.kaggle.com/c/jigsaw-unintended-bias-in-toxicity-classification}}. The original task is to classify the input sentence to one of five different labels for different degrees of toxicity, but we binarized the labels to only toxic and not toxic by merging all non-toxic examples. We down-sampled the dataset to 125k examples to accommodate for the available computational resources.
\end{enumerate}


An overview of each dataset is presented in Table \ref{tab:dataset_statistics}. Specifically, we show the size of each dataset in terms of the number of sentences in the training data, the number of occurrences of gender pronouns, and the percentage of positive (\textit{i.e.} sexist/toxic) examples. It is worth mentioning that gender words are detected using the \textit{gender-bender} Python package \footnote{\url{https://github.com/Garrett-R/gender_bender}}. An illustration of how our \textit{GE} score measures the importance of several counterfactual examples in terms of fairness from the Twitter and Wikipedia datasets is presented in Table \ref{tab:fairness_score_examples}.

\begin{table*}[h] 
\centering
\begin{tabular}{lllll}
\hline
 \textbf{Dataset} & \textbf{Size} & \textbf{Male pronouns} & \textbf{Female pronouns} & \textbf{Positives}\\
\hline
\centering

  Twitter     &     16,907 & 901  & 716  & 20.29\%       \\
  Wikipedia & 159,686  & 54,357 & 11,540  & 9.62\%      \\
  Jigsaw   & 125,000 & 51,134 & 13,937  & 5.98\%        \\     
\end{tabular}
\caption{Statistics of the three datasets used in our experiments. The male pronouns considered are \emph{he}, \emph{him}, \emph{himself}, and \emph{his}, while the female pronouns are \emph{she}, \emph{her}, \emph{hers}, and \emph{herself}.
} 
\label{tab:dataset_statistics}
\end{table*}

%

\begin{table*}[h] 
\centering
\begin{tabular}{lll}
\hline
\textbf{Factual}  & \textbf{Counterfactual}   & \textbf{\textit{GE}}  \\
\hline
\centering
okay king of the Wikipedia Nazis & okay queen of the Wikipedia Nazis & $0.36$\\
Kate you stupid woman!  & Kareem you stupid man! & 0.11\\
I'm not sexist.. But women drivers are terrible& I'm not sexist.. But men drivers are terrible& $0.10$\\
Oh my god.... When will this show end & Oh my god.... When will this show end   & $0.00$\\        
\end{tabular}
\caption{Examples of sentences from the Twitter and Wikipedia datasets with their \textit{GE} score show how important it is to include their counterfactual examples in the training data for bias mitigation. The \textit{GE} scores are obtained using BERT model.
}
\label{tab:fairness_score_examples}
\end{table*}


\subsection{Baselines}\label{baselines}
For relative comparison of our proposed method, we use (1) a vanilla model that is not trained using any bias mitigation heuristics, (2) counterfactual data augmentation (CDA) \cite{lu2020gender}, (3) counterfactual data substitution (CDS) \cite{maudslay2019s}. We also tried data balancing \cite{dixon2018measuring} and gender blindness \cite{garg2019counterfactual}, but we found them not effective for the specific datasets and models that we used. Similar to the work of \citet{maudslay2019s}, our implementation of CDA includes name flipping. 

\subsection{Evaluation metrics}\label{sec:FNED_FPED}
To measure fairness, we report the demographic parity (DP)  \cite{beutel2017data,hardt2016equality}, which is defined as:
 \begin{equation}
     DP = 1 - |p(\hat{y} = 1|z = 1) - p(\hat{y} = 1|z = 0)|
 \end{equation}
 where $\hat{y}$ refers to the model's prediction and $z$ $\in$ $\{0,1\}$ refers to keeping and flipping the gender words in the sentence, respectively. To compute the DP, we follow the procedure in other works \cite{dixon2018measuring,park2018reducing} that use a synthetic dataset called the Identity Phrase Templates Test Set (IPTTS) for measuring fairness metrics. We also use EqOdd and EqOpp1 as two additional fairness metrics, which we define in section A.2 in the technical appendix.  We use the area under the receiver operating characteristic curve (AUC) as our performance metric.


 

\begin{figure*}[h!]
     \centering
     \begin{subfigure}
    \centering
    \includegraphics[width=1\linewidth]{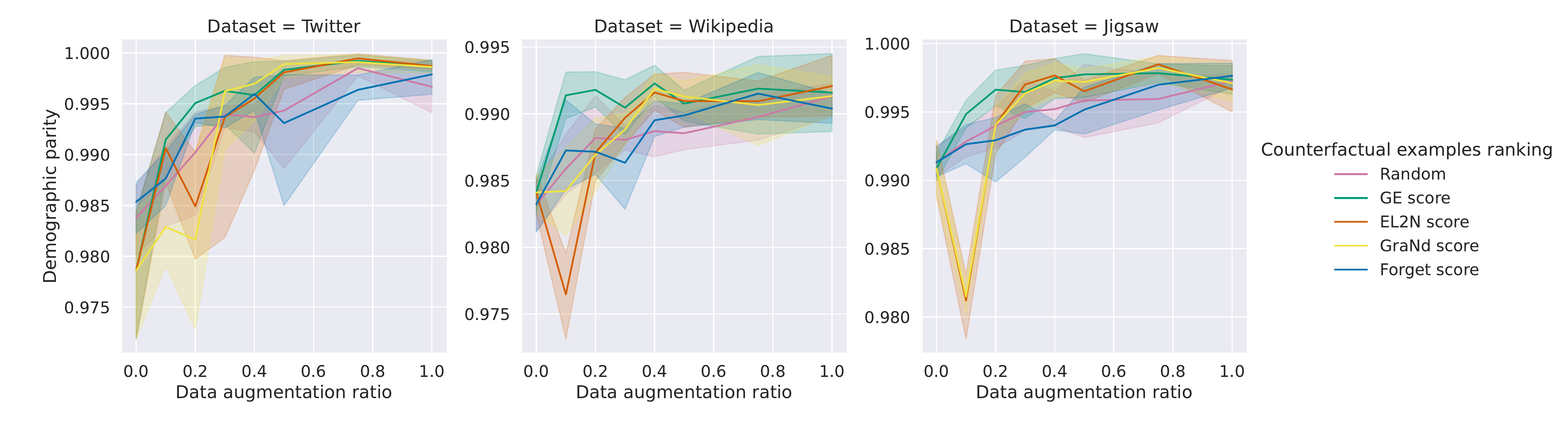}
     \end{subfigure}
     \begin{subfigure}
     \centering
    \includegraphics[width=1\linewidth]{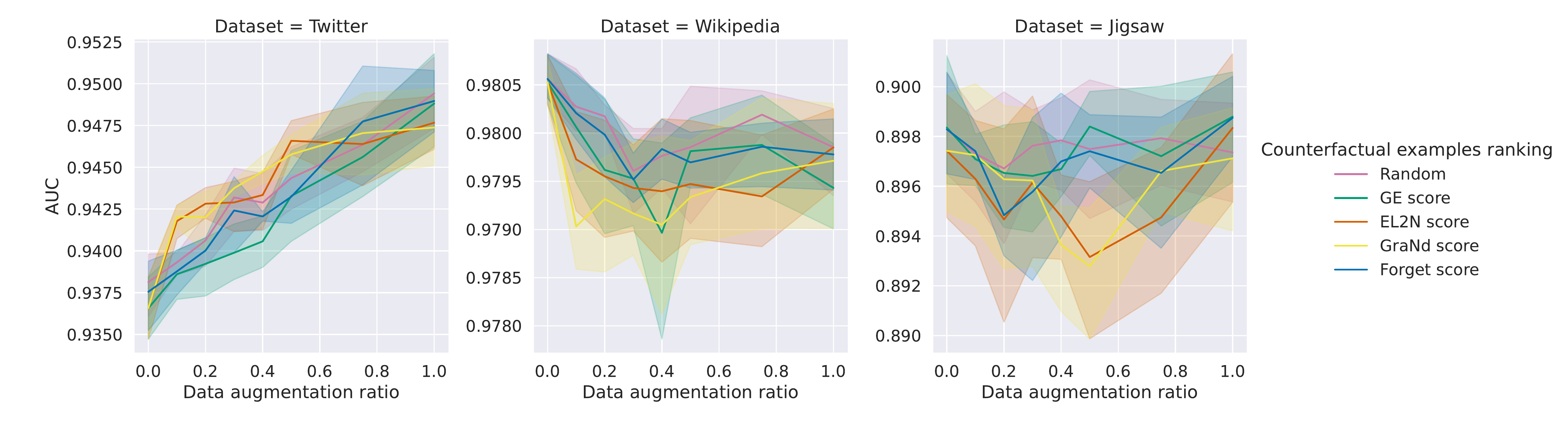}
     \end{subfigure}
        \caption{DP and AUC of BERT using different ratios and rankings of the counterfactual examples for data augmentation on the three datasets. The ratio represents the top $b$ $\%$ of the counterfactual examples. All plots are best viewed in color.}
        \label{fig:effect_of_important_examples_Bert}
\end{figure*}

\begin{figure*}[h!]
     \centering
     \begin{subfigure}
    \centering
    \includegraphics[width=1\linewidth]{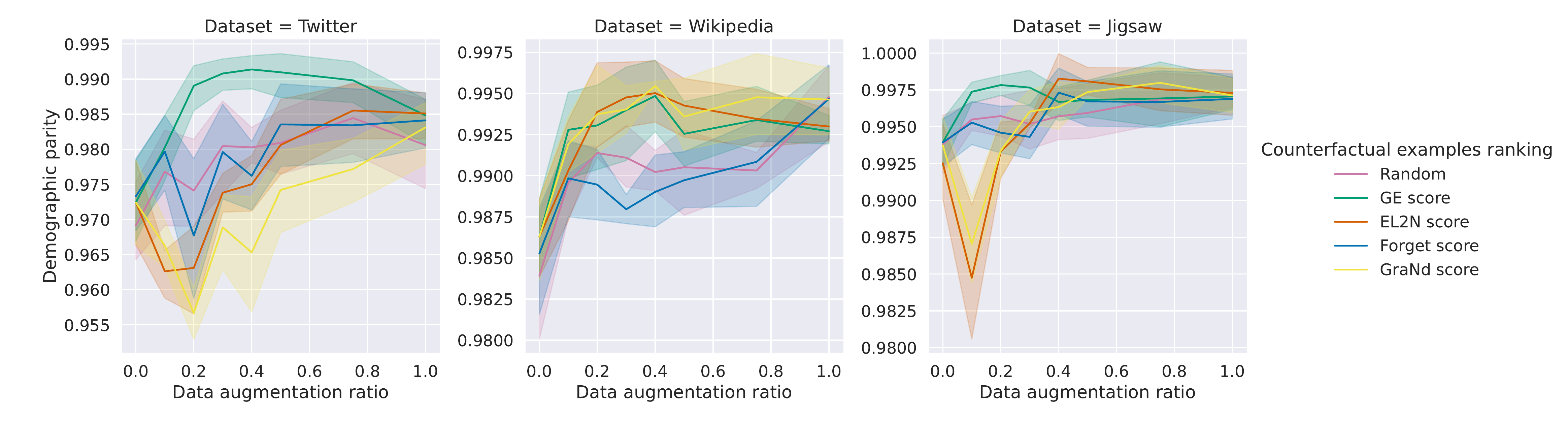}
     \end{subfigure}
     \begin{subfigure}
     \centering
    \includegraphics[width=1\linewidth]{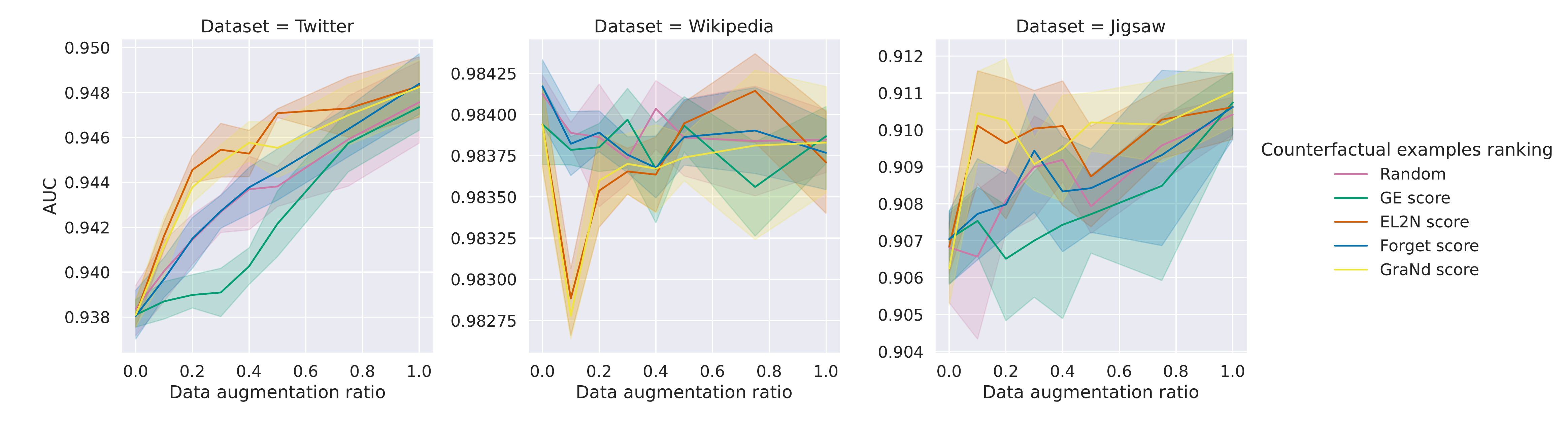}
     \end{subfigure}
        \caption{DP and AUC of RoBERTa using different ratios and rankings of the counterfactual examples for data augmentation on the three datasets. The ratio represents the top $b$ $\%$ of the counterfactual examples.
        }
        \label{fig:effect_of_important_examples_RoBERTa}
\end{figure*}
 
 \subsection{Experiment details}\label{exp_details}

We train the model for $15$ epochs using cross-entropy loss for both the Twitter and Wikipedia datasets, and $10$ epochs for the Jigsaw dataset. We used a ratio of $8$:$1$:$1$ between training, validation, and test for all datasets, except for the Wikipedia toxicity dataset, where we followed the split ratio used by \citet{dixon2018measuring}. For the \textit{GE} score, we chose the early stages of training to be the state of the model after one epoch. More details about the effect of selecting the epoch for calculating the \textit{GE} score are presented in section A.1 in the technical appendix. All the results are obtained by running the experiments for five different seeds. We used BERT and RoBERTa base models for text classification. Section A.3 in the technical appendix includes all the details needed about the hyper-parameters used to obtain the results. Our code will be publicly available for reproducibility \footnote{https://github.com/chandar-lab/healthy-data-diet}. Sections B.1, B.2, and B.3 in the code appendix provide more details about the code including the pre-processing of the datasets, the procedure to conduct and analyze the experiments, as well as the computing infrastructure used for running the experiments, respectively. 

\begin{table*}[h] 
\centering
\begin{tabular}{llllll}
\hline
\textbf{Ranking for the factuals} & \textbf{Ranking for the counterfactuals} & \textbf{Ranking name} \\
\hline
\centering
  \textcolor{black}{Ascending $GE$ score}  &  \textcolor{black}{Ascending $GE$ score}& \textcolor{black}{Vanilla GE}     \\ 
  Random  &  Random & Random     \\ 
  \textcolor{black}{Random}  &  \textcolor{black}{Descending \textit{GE}  score}& \textcolor{black}{Healthy random}     \\ 
  \textcolor{black}{Random} & \textcolor{black}{Ascending $GE$ score}  & \textcolor{black}{Unhealthy random}     \\

\end{tabular}
\caption{\textcolor{black}{The different ranking methods that we consider to prune the dataset in our experiments.}}
\label{tab:data_pruning_methods}
\end{table*}

\begin{figure*}[h!]
     \centering
     \begin{subfigure}
    \centering
    \includegraphics[width=1\linewidth, height = 2.6cm]{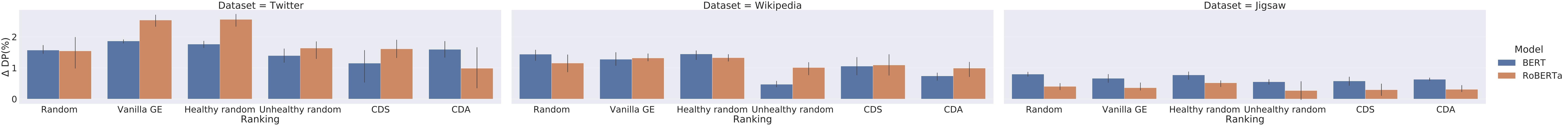}
     \end{subfigure}
     \begin{subfigure}
    \centering
    \includegraphics[width=1\linewidth, height = 2.6cm]{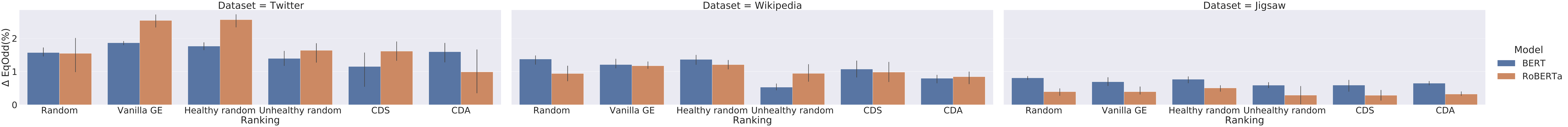}
     \end{subfigure}
     \begin{subfigure}
    \centering
    \includegraphics[width=1\linewidth, height = 2.6cm]{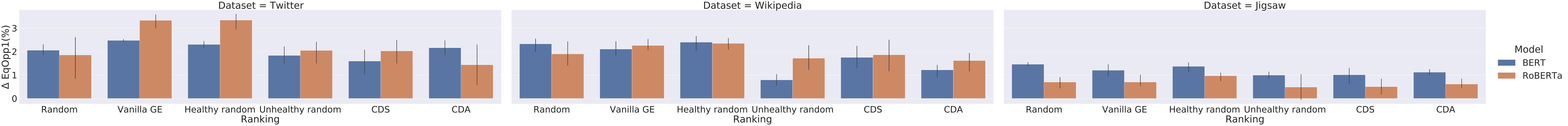}
     \end{subfigure}     
     \begin{subfigure}
     \centering
    \includegraphics[width=1\linewidth, height = 2.6cm]{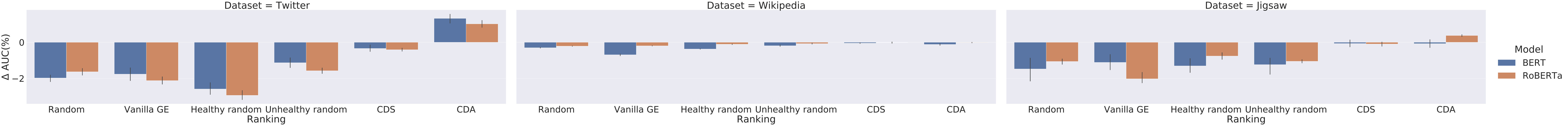}
     \end{subfigure}
        \caption{The percentage of change in the DP, EqOdd, EqOpp1, and AUC with BERT and RoBERTa on the three datasets using different pruning methods based on the ranking methods described in Table \ref{tab:data_pruning_methods}, compared to the biased model. Both CDA and CDS are added for comparison. For all the fairness metrics, higher values indicate fairer models. 
        }
        \label{fig:data_pruning_bert}
\end{figure*}

\paragraph{Experiment 1: Verifying that \textit{GE} score reflects the importance of counterfactual examples}

\noindent In this experiment, we want to verify empirically that our \textit{GE} score reflects the importance of each counterfactual example for fairness. Therefore, we train the model on all the factual data, as well as a varying ratio of the counterfactual examples. The counterfactual examples are ranked based on their \textit{GE} scores, such that we only use the top $b\%$ and prune the rest, where $b$ $\in$ $\{0,0.1,..,1\}$. We compared our \textit{GE} score with existing performance-based example scores, namely the (1) \textit{EL2N} score, (2) \textit{GraNd} score, (3) \textit{forget score}, as well as the (4) random score. Figs. \ref{fig:effect_of_important_examples_Bert}-\ref{fig:effect_of_important_examples_RoBERTa}
show that ranking the counterfactual sentences based on the \textit{GE} score results in a faster  increase in the DP, compared to other ranking scores. However, using RoBERTa on the Wikipedia dataset, both the \textit{EL2N} and \textit{GraNd} scores have almost the same rate of increase in DP as the \textit{GE} score. We hypothesize that this is due to a considerable overlap between the set of examples that are important for fairness (picked up by the \textit{GE} score) and the set of examples important for performance (picked up by the \textit{EL2N} and \textit{GraNd} scores), for this specific model and dataset. We verify and discuss our hypothesis in more detail in section A.4 in the technical appendix. Since we observe that the improvement in DP achieved by ranking the counterfactual examples based on the  \textit{GE} score is often accompanied by a degradation in the AUC, we explore the trade-off between fairness and performance in the next experiment.
\paragraph{Experiment 2: Finding the best trade-off between fairness and performance}

We choose $a$ and $b$ $\%$ of the samples from the factual and counterfactual examples, respectively, with $a$ $\in$ $\{0.3, 0.4, 0.5\}$ and $b$ $\in$ $\{0,0.1,0.2,0.3,0.4,0.5\}$. Table  \ref{tab:data_pruning_methods} shows the different settings that we consider to rank the factual and counterfactual examples. For every setting, Fig. \ref{fig:data_pruning_bert} shows the best configuration among all the different combinations of $a \%$ and $b \%$, based on the highest improvement in DP such that the degradation in AUC over the biased model is not more than $3\%$. It is clear that the model's fairness is affected by the ranking of the factual and counterfactual examples. When the factual examples are chosen randomly and the counterfactual examples are ranked based on the ascending \textit{GE} score, we get the worst fairness, which we refer to as the ``unhealthy random'' ranking. This occurs because we choose the least important counterfactual examples. The opposite happens when choosing the factual examples randomly and the counterfactual examples based on the descending \textit{GE} score, which we refer to as the ``healthy random'' ranking. This agrees with the intuition that provided in Section \ref{sec:combining_our_score_with_other_scores}.

The fairness obtained by our proposed healthy random ranking outperforms that of CDS and CDA on all the datasets with both BERT and RoBERTa models, using DP, EqOdd, and EqOpp1 metrics. This is accomplished with no more than $3$$\%$ degradation in the AUC on the downstream task, relative to the model without debiasing. 
Both the vanilla \textit{GE} and random rankings perform relatively well compared to CDA and CDS, but the healthy random ranking shows more improvement in terms of fairness. It is important to mention that all the ranking methods in Table \ref{tab:data_pruning_methods} use no more than half of the examples used by CDA. Note that the AUC for CDA is slightly higher than the baseline on the Twitter dataset because the dataset is related to sexism, so adding more counterfactual examples helps not only in improving fairness but also in improving the performance on the downstream task.

\section{Conclusion and Future Work}

The intuition behind the reduction in gender bias without a substantial degradation in the performance on different datasets stems from our hypothesis that models can find different ways, \textit{i.e.} different sets of weights, to solve the task. 
When the model is permitted to learn without any constraints being imposed, it tends to choose the simplest way to solve the given task, which includes learning some undesired associations between a group of features and the output labels (such as thinking that a man is more likely to be a doctor, than a woman). \textcolor{black}{Augmenting the training dataset with counterfactual examples during the fine-tuning step forces the model to \textit{unlearn} these undesired correlations and to find an alternative way to solve the task, which results in a less biased model. However, some of the examples in the augmented dataset can be redundant, or even have an adverse effect through enforcing gender stereotypes that the model had learned during pre-training. Our proposed method that ranks the examples based on the healthy random ranking defined in Table \ref{tab:data_pruning_methods} may then be used to prune such examples.}

Our future work will be in the direction of generalizing the applicability of our method, such that it includes the bias against other minority groups, such as LGBTQ+, Black people, Jewish people, etc. We believe that extending this work to other languages is also worth exploring. 
 
\section*{Acknowledgements} 

Sarath Chandar is supported by a Canada CIFAR AI Chair and an NSERC Discovery Grant. The authors acknowledge the computational resources provided by Microsoft. We are thankful to Nicolas Le Roux, Marc-Alexandre Côté, Eric Yuan, Andreas Madsen, Romain Laroche,  Su Lin Blodgett, and Adam Trischler for their helpful feedback in this project. We are also thankful to the reviewers for their constructive comments.
\newpage
\bibliography{bibliography}

\begin{thebibliography}{52}
\providecommand{\natexlab}[1]{#1}

\bibitem[{Adi et~al.(2017)Adi, Kermany, Belinkov, Lavi, and
  Goldberg}]{adi2016fine}
Adi, Y.; Kermany, E.; Belinkov, Y.; Lavi, O.; and Goldberg, Y. 2017.
\newblock Fine-grained analysis of sentence embeddings using auxiliary
  prediction tasks.
\newblock In \emph{International Conference on Learning Representations}.

\bibitem[{Belinkov and Glass(2019)}]{belinkov2019analysis}
Belinkov, Y.; and Glass, J. 2019.
\newblock Analysis methods in neural language processing: {A} survey.
\newblock \emph{Transactions of the Association for Computational Linguistics}.

\bibitem[{Beutel et~al.(2017)Beutel, Chen, Zhao, and Chi}]{beutel2017data}
Beutel, A.; Chen, J.; Zhao, Z.; and Chi, E.~H. 2017.
\newblock Data decisions and theoretical implications when adversarially
  learning fair representations.
\newblock \emph{Workshop on Fairness, Accountability, and Transparency in
  Machine Learning}.

\bibitem[{Brown et~al.(2020)Brown, Mann, Ryder, Subbiah, Kaplan, Dhariwal,
  Neelakantan, Shyam, Sastry, Askell et~al.}]{brown2020language}
Brown, T.; Mann, B.; Ryder, N.; Subbiah, M.; Kaplan, J.~D.; Dhariwal, P.;
  Neelakantan, A.; Shyam, P.; Sastry, G.; Askell, A.; et~al. 2020.
\newblock Language models are few-shot learners.
\newblock \emph{Advances in Neural Information Processing Systems}.

\bibitem[{Brunet et~al.(2019)Brunet, Alkalay-Houlihan, Anderson, and
  Zemel}]{brunet2019understanding}
Brunet, M.-E.; Alkalay-Houlihan, C.; Anderson, A.; and Zemel, R. 2019.
\newblock Understanding the origins of bias in word embeddings.
\newblock In \emph{International Conference on Machine Learning}.

\bibitem[{Budzianowski et~al.(2018)Budzianowski, Wen, Tseng, Casanueva, Ultes,
  Ramadan, and Ga{\v{s}}i{\'c}}]{budzianowski2018multiwoz}
Budzianowski, P.; Wen, T.-H.; Tseng, B.-H.; Casanueva, I.; Ultes, S.; Ramadan,
  O.; and Ga{\v{s}}i{\'c}, M. 2018.
\newblock {M}ulti{WOZ} - A Large-Scale Multi-Domain {W}izard-of-{O}z Dataset
  for Task-Oriented Dialogue Modelling.
\newblock In \emph{Conference on Empirical Methods in Natural Language
  Processing}.

\bibitem[{Chung et~al.(2014)Chung, Gulcehre, Cho, and
  Bengio}]{chung2014empirical}
Chung, J.; Gulcehre, C.; Cho, K.; and Bengio, Y. 2014.
\newblock Empirical evaluation of gated recurrent neural networks on sequence
  modeling.
\newblock In \emph{NeurIPS Workshop on Deep Learning}.

\bibitem[{Conneau et~al.(2018)Conneau, Kruszewski, Lample, Barrault, and
  Baroni}]{conneau2018you}
Conneau, A.; Kruszewski, G.; Lample, G.; Barrault, L.; and Baroni, M. 2018.
\newblock What you can cram into a single {\$}{\&}!{\#}* vector: Probing
  sentence embeddings for linguistic properties.
\newblock In \emph{Annual Meeting of the Association for Computational
  Linguistics}.

\bibitem[{De-Arteaga et~al.(2019)De-Arteaga, Romanov, Wallach, Chayes, Borgs,
  Chouldechova, Geyik, Kenthapadi, and Kalai}]{de2019bias}
De-Arteaga, M.; Romanov, A.; Wallach, H.; Chayes, J.; Borgs, C.; Chouldechova,
  A.; Geyik, S.; Kenthapadi, K.; and Kalai, A.~T. 2019.
\newblock Bias in bios: A case study of semantic representation bias in a
  high-stakes setting.
\newblock In \emph{Conference on Fairness, Accountability, and Transparency}.

\bibitem[{Devlin et~al.(2019)Devlin, Chang, Lee, and
  Toutanova}]{devlin2018bert}
Devlin, J.; Chang, M.; Lee, K.; and Toutanova, K. 2019.
\newblock {BERT: Pre-training of Deep Bidirectional Transformers for Language
  Understanding}.
\newblock In \emph{NAACL}, 4171--4186.

\bibitem[{Dinan et~al.(2020)Dinan, Fan, Wu, Weston, Kiela, and
  Williams}]{dinan2020multi}
Dinan, E.; Fan, A.; Wu, L.; Weston, J.; Kiela, D.; and Williams, A. 2020.
\newblock Multi-dimensional gender bias classification.
\newblock \emph{arXiv preprint arXiv:2005.00614}.

\bibitem[{Dixon et~al.(2018)Dixon, Li, Sorensen, Thain, and
  Vasserman}]{dixon2018measuring}
Dixon, L.; Li, J.; Sorensen, J.; Thain, N.; and Vasserman, L. 2018.
\newblock Measuring and mitigating unintended bias in text classification.
\newblock In \emph{Conference on AI, Ethics, and Society}.

\bibitem[{Friedman and Nissenbaum(1996)}]{friedman1996bias}
Friedman, B.; and Nissenbaum, H. 1996.
\newblock Bias in computer systems.
\newblock \emph{Transactions on Information Systems}.

\bibitem[{Garg et~al.(2019)Garg, Perot, Limtiaco, Taly, Chi, and
  Beutel}]{garg2019counterfactual}
Garg, S.; Perot, V.; Limtiaco, N.; Taly, A.; Chi, E.~H.; and Beutel, A. 2019.
\newblock Counterfactual fairness in text classification through robustness.
\newblock In \emph{Conference on AI, Ethics, and Society}.

\bibitem[{Gupta et~al.(2021)Gupta, Ferber, Dilkina, and
  Ver~Steeg}]{gupta2021controllable}
Gupta, U.; Ferber, A.; Dilkina, B.; and Ver~Steeg, G. 2021.
\newblock Controllable Guarantees for Fair Outcomes via Contrastive Information
  Estimation.
\newblock In \emph{AAAI Conference on Artificial Intelligence}.

\bibitem[{Hall~Maudslay et~al.(2019)Hall~Maudslay, Gonen, Cotterell, and
  Teufel}]{maudslay2019s}
Hall~Maudslay, R.; Gonen, H.; Cotterell, R.; and Teufel, S. 2019.
\newblock It{'}s All in the Name: Mitigating Gender Bias with Name-Based
  Counterfactual Data Substitution.
\newblock In \emph{Conference on Empirical Methods in Natural Language
  Processing and International Joint Conference on Natural Language
  Processing}.

\bibitem[{Hardt, Price, and Srebro(2016)}]{hardt2016equality}
Hardt, M.; Price, E.; and Srebro, N. 2016.
\newblock Equality of opportunity in supervised learning.
\newblock \emph{Advances in Neural Information Processing Systems}.

\bibitem[{Hearst et~al.(1998)Hearst, Dumais, Osuna, Platt, and
  Scholkopf}]{hearst1998support}
Hearst, M.~A.; Dumais, S.~T.; Osuna, E.; Platt, J.; and Scholkopf, B. 1998.
\newblock Support vector machines.
\newblock \emph{Intelligent Systems and their applications}.

\bibitem[{Hochreiter and Schmidhuber(1997)}]{hochreiter1997long}
Hochreiter, S.; and Schmidhuber, J. 1997.
\newblock {Long Short-Term Memory}.
\newblock \emph{Neural Computation}, 9(8): 1735--1780.

\bibitem[{Hupkes, Veldhoen, and Zuidema(2018)}]{hupkes2018visualisation}
Hupkes, D.; Veldhoen, S.; and Zuidema, W. 2018.
\newblock Visualisation and 'diagnostic classifiers' reveal how recurrent and
  recursive neural networks process hierarchical structure.
\newblock \emph{Journal of Artificial Intelligence Research}.

\bibitem[{Isabelle, Cherry, and Foster(2017)}]{isabelle2017challenge}
Isabelle, P.; Cherry, C.; and Foster, G. 2017.
\newblock A Challenge Set Approach to Evaluating Machine Translation.
\newblock In \emph{Conference on Empirical Methods in Natural Language
  Processing}.

\bibitem[{Jaiswal et~al.(2020)Jaiswal, Moyer, Ver~Steeg, AbdAlmageed, and
  Natarajan}]{jaiswal2020invariant}
Jaiswal, A.; Moyer, D.; Ver~Steeg, G.; AbdAlmageed, W.; and Natarajan, P. 2020.
\newblock Invariant representations through adversarial forgetting.
\newblock In \emph{AAAI Conference on Artificial Intelligence}.

\bibitem[{Kiritchenko and Mohammad(2018)}]{kiritchenko2018examining}
Kiritchenko, S.; and Mohammad, S. 2018.
\newblock Examining Gender and Race Bias in Two Hundred Sentiment Analysis
  Systems.
\newblock In \emph{Joint Conference on Lexical and Computational Semantics}.

\bibitem[{Liu et~al.(2019)Liu, Ott, Goyal, Du, Joshi, Chen, Levy, Lewis,
  Zettlemoyer, and Stoyanov}]{liu2019roberta}
Liu, Y.; Ott, M.; Goyal, N.; Du, J.; Joshi, M.; Chen, D.; Levy, O.; Lewis, M.;
  Zettlemoyer, L.; and Stoyanov, V. 2019.
\newblock {RoBERTa}: {A} Robustly Optimized {BERT} Pretraining Approach.
\newblock \emph{arXiv preprint arXiv:1907.11692}.

\bibitem[{Lu et~al.(2020)Lu, Mardziel, Wu, Amancharla, and
  Datta}]{lu2020gender}
Lu, K.; Mardziel, P.; Wu, F.; Amancharla, P.; and Datta, A. 2020.
\newblock Gender bias in neural natural language processing.
\newblock In \emph{Logic, Language, and Security}.

\bibitem[{Madras et~al.(2018)Madras, Creager, Pitassi, and
  Zemel}]{madras2018learning}
Madras, D.; Creager, E.; Pitassi, T.; and Zemel, R. 2018.
\newblock Learning adversarially fair and transferable representations.
\newblock In \emph{International Conference on Machine Learning}.

\bibitem[{Manzini et~al.(2019)Manzini, Yao~Chong, Black, and
  Tsvetkov}]{manzini2019black}
Manzini, T.; Yao~Chong, L.; Black, A.~W.; and Tsvetkov, Y. 2019.
\newblock Black is to Criminal as Caucasian is to Police: Detecting and
  Removing Multiclass Bias in Word Embeddings.
\newblock In \emph{Conference of the North {A}merican Chapter of the
  Association for Computational Linguistics}.

\bibitem[{May et~al.(2019)May, Wang, Bordia, Bowman, and
  Rudinger}]{may2019measuring}
May, C.; Wang, A.; Bordia, S.; Bowman, S.~R.; and Rudinger, R. 2019.
\newblock On Measuring Social Biases in Sentence Encoders.
\newblock In \emph{Conference of the North {A}merican Chapter of the
  Association for Computational Linguistics}.

\bibitem[{Meade, Poole-Dayan, and Reddy(2022)}]{meade2021empirical}
Meade, N.; Poole-Dayan, E.; and Reddy, S. 2022.
\newblock An Empirical Survey of the Effectiveness of Debiasing Techniques for
  Pre-trained Language Models.
\newblock In \emph{{Annual} {Meeting} of the {Association} for {Computational}
  {Linguistics}}.

\bibitem[{Mordido and Meinel(2020)}]{mark-evaluate}
Mordido, G.; and Meinel, C. 2020.
\newblock Mark-Evaluate: Assessing Language Generation using Population
  Estimation Methods.
\newblock In \emph{International Conference on Computational Linguistics}.

\bibitem[{Nadeem, Bethke, and Reddy(2021)}]{nadeem2020stereoset}
Nadeem, M.; Bethke, A.; and Reddy, S. 2021.
\newblock {S}tereo{S}et: Measuring stereotypical bias in pretrained language
  models.
\newblock In \emph{Annual Meeting of the Association for Computational
  Linguistics and International Joint Conference on Natural Language
  Processing}.

\bibitem[{Naik et~al.(2018)Naik, Ravichander, Sadeh, Rose, and
  Neubig}]{naik2018stress}
Naik, A.; Ravichander, A.; Sadeh, N.; Rose, C.; and Neubig, G. 2018.
\newblock Stress Test Evaluation for Natural Language Inference.
\newblock In \emph{International Conference on Computational Linguistics}.

\bibitem[{Park, Shin, and Fung(2018)}]{park2018reducing}
Park, J.~H.; Shin, J.; and Fung, P. 2018.
\newblock Reducing Gender Bias in Abusive Language Detection.
\newblock In \emph{Conference on Empirical Methods in Natural Language
  Processing}.

\bibitem[{Paul, Ganguli, and Dziugaite(2021)}]{paul2021deep}
Paul, M.; Ganguli, S.; and Dziugaite, G.~K. 2021.
\newblock Deep Learning on a Data Diet: Finding Important Examples Early in
  Training.
\newblock \emph{Advances in Neural Information Processing Systems}.

\bibitem[{Pearl(1995)}]{pearl1995}
Pearl, J. 1995.
\newblock Causal Diagrams for Empirical Research.
\newblock \emph{Biometrika}.

\bibitem[{Radford et~al.(2018)Radford, Narasimhan, Salimans, and
  Sutskever}]{radford2018improving}
Radford, A.; Narasimhan, K.; Salimans, T.; and Sutskever, I. 2018.
\newblock Improving language understanding by generative pre-training.
\newblock \emph{OpenAI report}.

\bibitem[{Radford et~al.(2019)Radford, Wu, Child, Luan, Amodei, and
  Sutskever}]{radford2019language}
Radford, A.; Wu, J.; Child, R.; Luan, D.; Amodei, D.; and Sutskever, I. 2019.
\newblock {Language Models are Unsupervised Multitask Learners}.
\newblock \emph{OpenAI Blog}, 1(8): 9.

\bibitem[{Raffel et~al.(2020)Raffel, Shazeer, Roberts, Lee, Narang, Matena,
  Zhou, Li, and Liu}]{raffel2019exploring}
Raffel, C.; Shazeer, N.; Roberts, A.; Lee, K.; Narang, S.; Matena, M.; Zhou,
  Y.; Li, W.; and Liu, P.~J. 2020.
\newblock Exploring the Limits of Transfer Learning with a Unified Text-to-Text
  Transformer.
\newblock \emph{Journal of Machine Learning Research}.

\bibitem[{Rajpurkar et~al.(2016)Rajpurkar, Zhang, Lopyrev, and
  Liang}]{rajpurkar2016squad}
Rajpurkar, P.; Zhang, J.; Lopyrev, K.; and Liang, P. 2016.
\newblock {SQ}u{AD}: 100,000+ Questions for Machine Comprehension of Text.
\newblock In \emph{Conference on Empirical Methods in Natural Language
  Processing}.

\bibitem[{Sauder et~al.(2020)Sauder, Hu, Che, Mordido, Yang, and
  Meinel}]{sauder-etal-2020-best}
Sauder, J.; Hu, T.; Che, X.; Mordido, G.; Yang, H.; and Meinel, C. 2020.
\newblock Best Student Forcing: A Simple Training Mechanism in Adversarial
  Language Generation.
\newblock In \emph{Language Resources and Evaluation Conference}.

\bibitem[{Sennrich(2017)}]{sennrich2016grammatical}
Sennrich, R. 2017.
\newblock How Grammatical is Character-level Neural Machine Translation?
  {A}ssessing {MT} Quality with Contrastive Translation Pairs.
\newblock In \emph{{E}uropean Chapter of the Association for Computational
  Linguistics}.

\bibitem[{Song et~al.(2019)Song, Kalluri, Grover, Zhao, and
  Ermon}]{song2019learning}
Song, J.; Kalluri, P.; Grover, A.; Zhao, S.; and Ermon, S. 2019.
\newblock Learning controllable fair representations.
\newblock In \emph{International Conference on Artificial Intelligence and
  Statistics}.

\bibitem[{Tenney, Das, and Pavlick(2019)}]{tenney2019bert}
Tenney, I.; Das, D.; and Pavlick, E. 2019.
\newblock {BERT} Rediscovers the Classical {NLP} Pipeline.
\newblock In \emph{Annual Meeting of the Association for Computational
  Linguistics}.

\bibitem[{Toneva et~al.(2019)Toneva, Sordoni, des Combes, Trischler, Bengio,
  and Gordon}]{toneva2018empirical}
Toneva, M.; Sordoni, A.; des Combes, R.~T.; Trischler, A.; Bengio, Y.; and
  Gordon, G.~J. 2019.
\newblock An Empirical Study of Example Forgetting during Deep Neural Network
  Learning.
\newblock In \emph{International Conference on Learning Representations}.

\bibitem[{Vaswani et~al.(2017)Vaswani, Shazeer, Parmar, Uszkoreit, Jones,
  Gomez, Kaiser, and Polosukhin}]{vaswani2017attention}
Vaswani, A.; Shazeer, N.; Parmar, N.; Uszkoreit, J.; Jones, L.; Gomez, A.~N.;
  Kaiser, {\L}.; and Polosukhin, I. 2017.
\newblock {Attention is All You Need}.
\newblock In \emph{{Proc. of the Advances in Neural Information Processing
  Systems (Neurips)}}, 5998--6008.

\bibitem[{Vig et~al.(2020)Vig, Gehrmann, Belinkov, Qian, Nevo, Singer, and
  Shieber}]{vig2020investigating}
Vig, J.; Gehrmann, S.; Belinkov, Y.; Qian, S.; Nevo, D.; Singer, Y.; and
  Shieber, S. 2020.
\newblock Investigating Gender Bias in Language Models Using Causal Mediation
  Analysis.
\newblock In \emph{Advances in Neural Information Processing Systems}.

\bibitem[{Wang et~al.(2019)Wang, Pruksachatkun, Nangia, Singh, Michael, Hill,
  Levy, and Bowman}]{wang2019superglue}
Wang, A.; Pruksachatkun, Y.; Nangia, N.; Singh, A.; Michael, J.; Hill, F.;
  Levy, O.; and Bowman, S. 2019.
\newblock Super{GLUE}: A stickier benchmark for general-purpose language
  understanding systems.
\newblock \emph{Advances in Neural Information Processing Systems}.

\bibitem[{Wang et~al.(2018)Wang, Singh, Michael, Hill, Levy, and
  Bowman}]{wang2018glue}
Wang, A.; Singh, A.; Michael, J.; Hill, F.; Levy, O.; and Bowman, S. 2018.
\newblock {GLUE}: A Multi-Task Benchmark and Analysis Platform for Natural
  Language Understanding.
\newblock In \emph{{EMNLP} Workshop {B}lackbox{NLP}: Analyzing and Interpreting
  Neural Networks for {NLP}}.

\bibitem[{Waseem and Hovy(2016)}]{waseem2016hateful}
Waseem, Z.; and Hovy, D. 2016.
\newblock Hateful Symbols or Hateful People? {P}redictive Features for Hate
  Speech Detection on {T}witter.
\newblock In \emph{{NAACL} Student Research Workshop}.

\bibitem[{Yang et~al.(2019)Yang, Dai, Yang, Carbonell, Salakhutdinov, and
  Le}]{yang2019xlnet}
Yang, Z.; Dai, Z.; Yang, Y.; Carbonell, J.; Salakhutdinov, R.~R.; and Le, Q.~V.
  2019.
\newblock {XLNet}: {G}eneralized autoregressive pretraining for language
  understanding.
\newblock \emph{Advances in Neural Information Processing Systems}.

\bibitem[{Zhang et~al.(2020)Zhang, Bai, Zhang, Bai, Zhu, and
  Zhao}]{zhang2020demographics}
Zhang, G.; Bai, B.; Zhang, J.; Bai, K.; Zhu, C.; and Zhao, T. 2020.
\newblock Demographics Should Not Be the Reason of Toxicity: Mitigating
  Discrimination in Text Classifications with Instance Weighting.
\newblock In \emph{Annual Meeting of the Association for Computational
  Linguistics}.

\bibitem[{Zhang et~al.(2018)Zhang, Dinan, Urbanek, Szlam, Kiela, and
  Weston}]{zhang2018personalizing}
Zhang, S.; Dinan, E.; Urbanek, J.; Szlam, A.; Kiela, D.; and Weston, J. 2018.
\newblock Personalizing Dialogue Agents: {I} have a dog, do you have pets too?
\newblock In \emph{Annual Meeting of the Association for Computational
  Linguistics}.

\end{thebibliography}
\newpage
\appendix
\section{Technical appendix}
\subsection{Robustness of the \textit{GE} score}
As mentioned in Section 4.1 in the main paper, we consider the \textit{GE} score to be both dataset and model-dependent, and it is computed after $1$ epoch of the model training. In this section, we study the effect of changing the dataset, model architecture and initialization, as well as the number of epochs after which we compute the \textit{GE} score.

\subsubsection{The effect of changing the model initialization on the \textit{GE} score}
\begin{figure}[h!]
     \centering
    \centering
    \includegraphics[width=1\linewidth,height = 3cm]{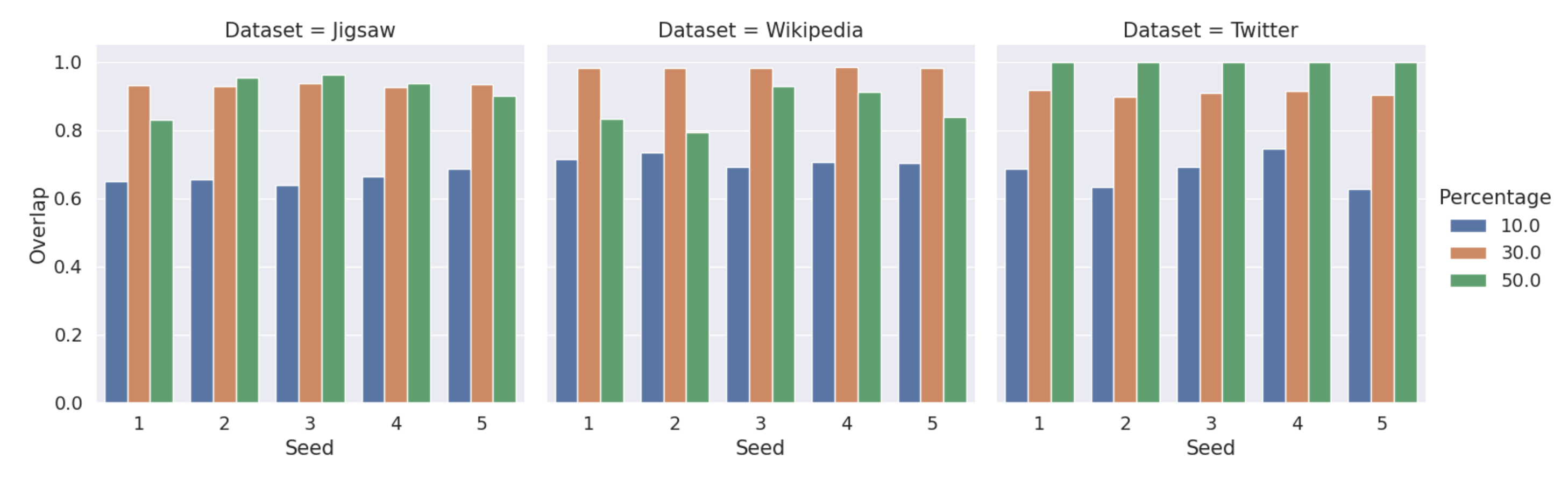}
        \caption{The ratio of overlap between the examples that we select based on the average \textit{GE} score, and the ones we select based on the individual \textit{GE} scores from different seeds. The overlap is computed for different percentages of examples. The experiments were conducted using BERT on all three datasets.
        }
        \label{fig:scores_overlap_different_seeds}
\end{figure}

The \textit{GE} score is computed as the average score over 5 different seeds. To understand the extent to which the scores can deviate by changing the seed, we study the agreement between the  \textit{GE} score, averaged over 5 different seeds, and the \textit{GE} score obtained from each seed. Fig. \ref{fig:scores_overlap_different_seeds} shows the ratio of overlap between the examples picked up based on the average score and those picked up based on the individual scores from all the seeds. The experiments are conducted using BERT on all three datasets. The observation is that different initializations yield \textit{GE} scores that agree to a large extent, with more than $60$$\%$ overlap when we only choose $10$$\%$ of the examples and more than $80$$\%$ overlap when choosing $30$$\%$ or more.

\subsubsection{The effect of changing the number of early epochs on the \textit{GE} score}
\begin{figure}[h!]
     \centering
    \centering
    \includegraphics[width=1\linewidth, height = 3cm]{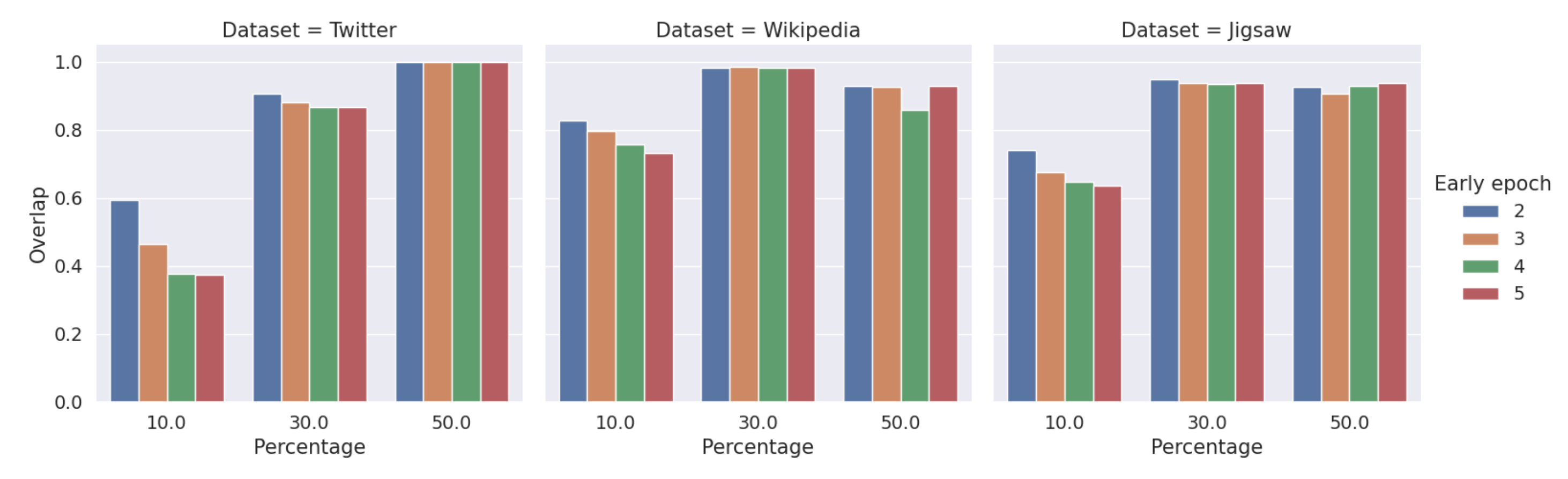}
        \caption{The ratio of overlap between the examples selected according to the 
        \textit{GE} score when the number of early epochs is set to $1$ (our choice in the main paper) and when it is set to $2$, $3$, $4$, and $5$ epochs. The overlap is computed for different percentages of the examples. The number of early epochs is the number of epochs after which the  \textit{GE} score is computed. The experiments were conducted using BERT on all three datasets.
        }
        \label{fig:scores_overlap_early_epochs}
\end{figure}

Our experiments in the main paper were conducted using the \textit{GE} score computed after $1$ epoch of the model training, which we refer to as the stage of early training. To explore the effect of changing the number of epochs in the early training, we study the agreement between the \textit{GE} score obtained after $1$ epoch, and that obtained after $2$, $3$, $4$, and $5$ epochs. Fig. \ref{fig:scores_overlap_early_epochs} shows the ratio of overlap between the examples selected according to the \textit{GE} score when the number of early epochs is set to $1$ and when it is set to $2$, $3$, $4$, and $5$ epochs. We observe more than $80$$\%$ overlap when selecting $30$$\%$ of the examples or more.

\subsubsection{The effect of changing the model architecture on the \textit{GE} score}
\begin{figure}[h!]
     \centering
    \centering
    \includegraphics[width=0.6\linewidth,height = 3cm]{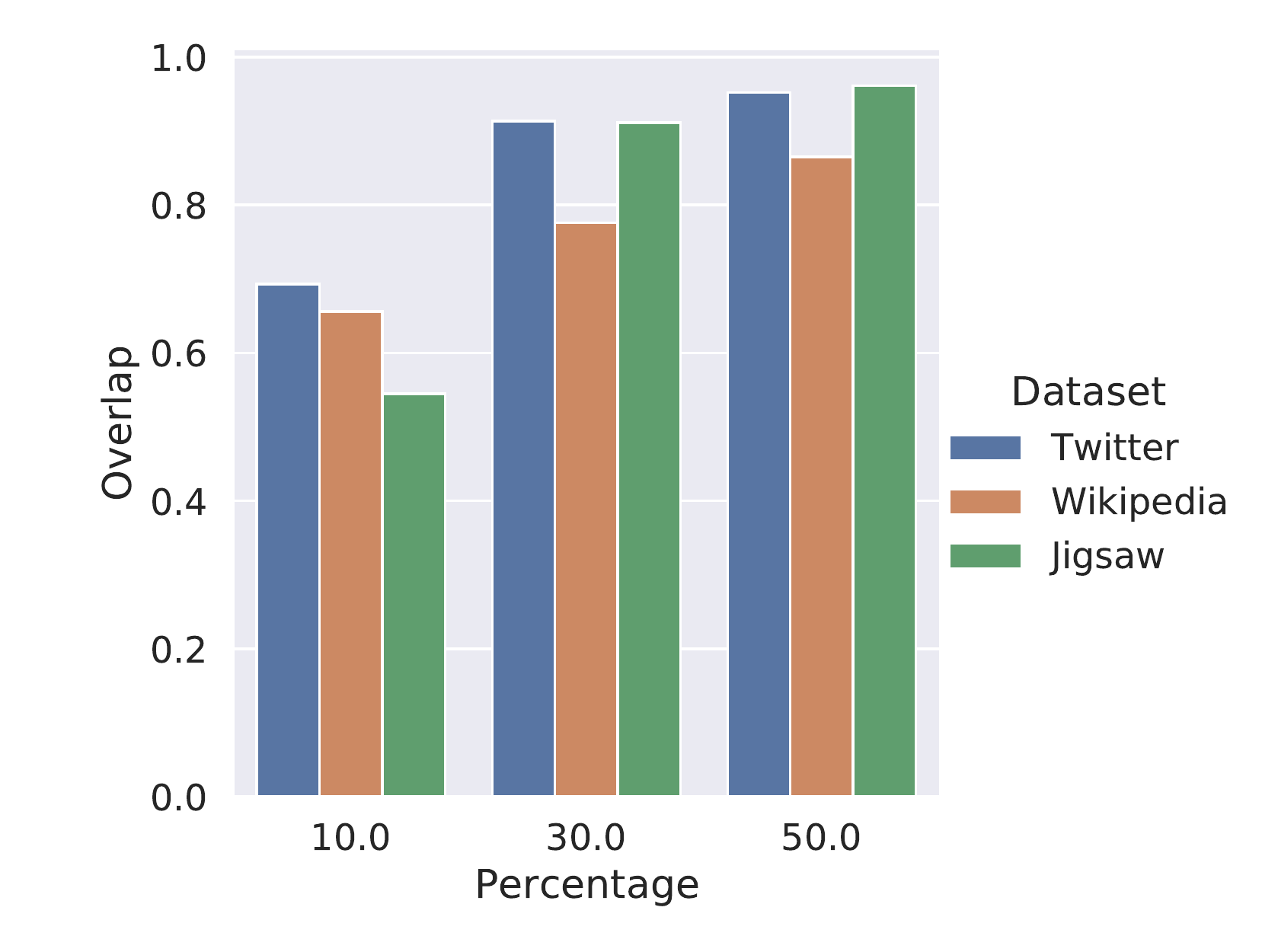}
        \caption{The ratio of overlap between the selected examples based on the \textit{GE} score when the score is computed separately for BERT and RoBERTa models. The overlap is computed for different percentages of the examples. The experiments were conducted on all the datasets.
        }
        \label{fig:scores_overlap_different_models}
\end{figure}

To study the effect of changing the model architecture on the \textit{GE} score, we compute it for both BERT and RoBERTa separately as shown in Eq. 2. We then pick up groups of examples based on both scores, such that the percentage of examples is the same in the groups. We then vary the percentage and repeat the experiment for all datasets. The results in Fig. \ref{fig:scores_overlap_different_models} suggest that the examples in both groups have considerable overlap. We consider the \textit{GE} score to be model-dependent for generality. 

\subsection{Additional fairness metrics}
In this section, we define the equality of opportunity and equality of odds metrics, which are used to assess the model's fairness in our experiments, as follows:
\begin{itemize}
\item Equality of opportunity for $y=1$ (EqOpp1):  
 \begin{equation}
     1 - |p(\hat{y} = 1|z = 1, y=1) - p(\hat{y} = 1|z = 0, y=1)|
 \end{equation}
 
 \item Equality of opportunity for $y=0$ (EqOpp0):  
 \begin{equation}
     1 - |p(\hat{y} = 1|z = 1, y=0) - p(\hat{y} = 1|z = 0, y=0)|
 \end{equation}
 
 \item Equality of odds (EqOdd):  
 \begin{equation}
     0.5 \times (EqOpp1 + EqOpp0)
 \end{equation}
\end{itemize}
 where $\hat{y}$ refers to the model's prediction, $y$ refers to the ground truth label, and $z$ $\in$ $\{0,1\}$ refers to the gender words in the sentence before and after flipping, respectively.

\subsection{Hyper-parameter tuning}
In this section, we list all the hyper-parameters used in the experiments in Tables \ref{tab:hyperparamaters}-\ref{tab:a_b_combinations}, along with the range of values for each one and their final values. The choice is based on the performance on the validation dataset.

\begin{table*}[h] 
\centering
\begin{tabular}{llllllll}
\hline
\textbf{Hyper-parameter} & \textbf{Values tried} & \textbf{Value used}\\
\hline
\centering

         Learning rate                      &  \{$10^{-2}$,$10^{-3}$,$10^{-4}$,$10^{-5}$,$10^{-6}$,$10^{-7}$\}    &$10^{-6}$ &\\
         Batch size (for Twitter dataset)     &  \{$16$, $32$, $64$, $128$\}    &$64$ &\\
         Batch size (for Wikipedia dataset)     &  \{$16$, $32$, $64$, $128$\}  &$32$ &\\
         Batch size (for Jigsaw dataset)     &  \{$16$, $32$, $64$, $128$\}    &$32$ &\\
         Max length (for the Twitter and Wikipedia dataset)     &  Not set (default value)    &Not set (default value) &\\
         Max length (for Jigsaw dataset)     &  \{$40$, $60$, $80$, $100$\}    &$40$ &\\
         The ratio of factual examples ($a$)&  \{$0.3$, $0.4$, $0.5$\}    & see Table \ref{tab:a_b_combinations} &\\
         The ratio of counterfactual examples ($b$)&  \{$0$, $0.1$, $0.2$, $0.3$, $0.4$, $0.5$\}    & see Table \ref{tab:a_b_combinations} &\\
\end{tabular}
\caption{This table shows the list of hyper-parameters used in the experiments, the range of values tried for each, and the final values.
}
\label{tab:hyperparamaters}
\end{table*}

\begin{table*}[h] 
\centering
\begin{tabular}{llllllll}
\hline
\textbf{Dataset} & \textbf{Model} & \textbf{Ranking} & \textbf{a (\%)}& \textbf{b (\%)}\\
\hline
\centering

                                       && Healthy random        &  $40$     &$40$   \\ 
       &\textsc{BERT}                  & Unhealthy random       &  $40$     &$50$   \\ 
                                       && Vanilla GE            &  $50$     &$50$   \\ 
                                       && Random                &  $40$     &$40$   \\ 
Twitter                            \\
        
                                       && Healthy random        &  $30$     &$40$   \\ 
       &Ro\textsc{BERT}a                 & Unhealthy random      &  $40$     &$40$   \\ 
                                       && Vanilla GE            &  $40$     &$50$   \\ 
                                       && Random                &  $40$     &$40$   \\ 
\hline
                                       && Healthy random        &  $30$     &$10$   \\ 
       &\textsc{BERT}                  & Unhealthy random       &  $50$     &$50$   \\ 
                                       && Vanilla GE            &  $30$     &$10$   \\ 
                                       && Random                &  $30$     &$40$   \\ 
Wikipedia                            \\
        
                                       && Healthy random        &  $30$     &$30$   \\ 
       &Ro\textsc{BERT}a                 & Unhealthy random      &  $40$     &$50$   \\ 
                                       && Vanilla GE            &  $30$     &$10$   \\ 
                                       && Random                &  $40$     &$30$   \\ 
\hline
                                       && Healthy random        &  $40$     &$50$   \\ 
       &\textsc{BERT}                  & Unhealthy random       &  $30$     &$50$   \\ 
                                       && Vanilla GE            &  $50$     &$50$   \\ 
                                       && Random                &  $40$     &$50$   \\ 
Jigsaw                            \\
        
                                       && Healthy random        &  $40$     &$50$   \\ 
       &Ro\textsc{BERT}a                 & Unhealthy random      &  $30$     &$50$   \\ 
                                       && Vanilla GE            &  $30$     &$30$   \\ 
                                       && Random                &  $40$     &$50$   \\ 
\end{tabular}
\caption{This table shows the combinations of the ratios of examples from the factual and counterfactual examples that yield the highest improvement in terms of demographic parity such that the degradation in AUC is not more than $3$$\%$, compared to the biased model. The results are shown for different datasets, models, and rankings according to Table \ref{tab:data_pruning_methods}.
}
\label{tab:a_b_combinations}
\end{table*}

\subsection{Ablation Studies}
To have a better understanding of our results, we also conduct an ablation study to answer the question that we posed in Section \ref{exp_details}.

\subsubsection{Why does adding more counterfactual examples that are ranked according to the $EL2N$ and $GraNd$ scores result in the same rate of increase in demographic parity as ranking them according to \textit{GE} score, on the Wikipedia dataset using RoBERTa model?}

\begin{figure}
     \centering
    \centering
    \includegraphics[width=1\linewidth,height = 3cm]{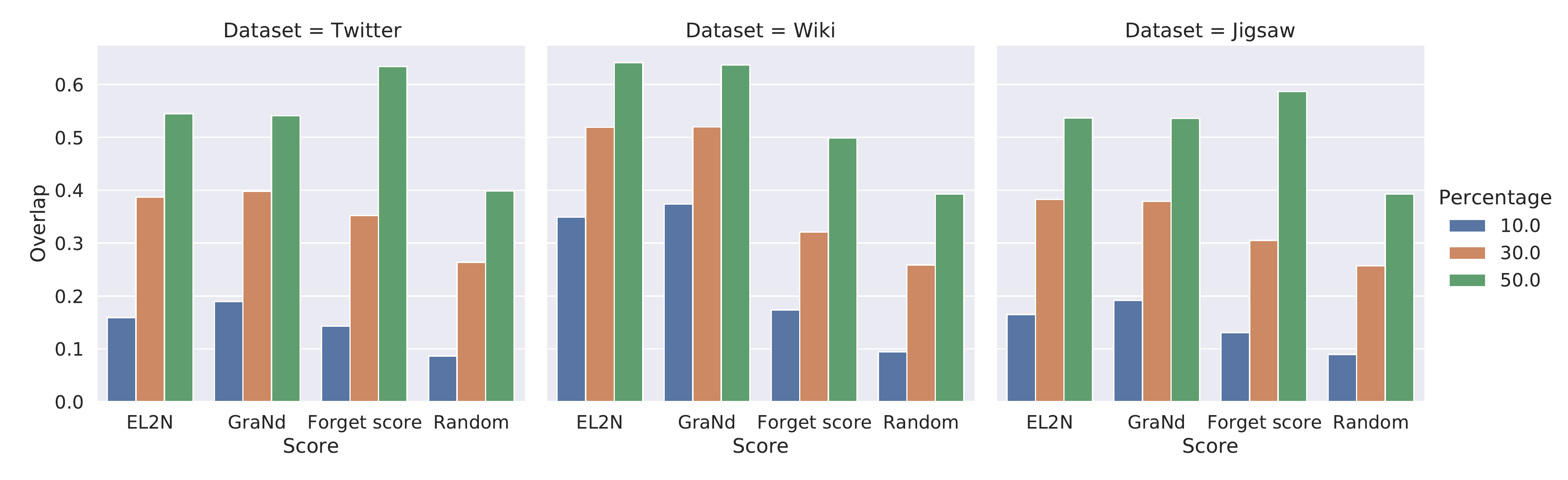}
        \caption{The ratio of overlap between the examples chosen based on the \textit{GE} score and the ones chosen based on other scores, for different percentages of the examples. The experiments were conducted using RoBERTa on all three datasets.
        }
        \label{fig:scores_overlap_Roberta}
\end{figure}

The results of experiment 1 in Fig. 3 in the main paper suggest that both $EL2N$ and $GraNd$ yield almost the same rate of increase in demographic parity as the \textit{GE} score with the same ratio of the counterfactual examples on the Wikipedia dataset and using RoBERTa model. We mentioned in the paper that this is attributed to having a considerable overlap between the set of examples that are important for fairness (picked up by the \textit{GE} score) and the set of examples important for performance (picked up by the \textit{EL2N} and \textit{GraNd} scores) on the specific dataset and model that we used. To make our argument convincing, we show the ratio of overlap between the examples that we choose from the \textit{GE} score and the ones we choose from all other scores for different datasets using RoBERTa model in Fig. \ref{fig:scores_overlap_Roberta}. For different percentages of the chosen examples, we find that both $EL2N$ and $GraNd$ have the highest overlap with the \textit{GE} score on the Wikipedia dataset.


\subsection{Ethical and Societal Consequences}
Although we focused on gender bias, our work may be generalized to mitigate the model's bias against any other group. We hope that this can be a step towards having models that can be safely deployed in real life without having ethical or societal concerns.

\section{Code appendix}
\subsection{Dataset pre-processing}

The data pre-pre-processing can be summarized in the following points:
\begin{itemize}
    \item Filling the rows that have NAN values.
    \item Binarizing the labels. In the Twitter dataset, the original labels are sexist, racist, and neither sexist nor racist. We binarized them into sexist and not sexist. For the Wikipedia and Jigsaw datasets, we convert the degree of toxicity, which goes from $0$ to $1$, into a binary label by setting a threshold at $0.5$
    \item Creating the counterfactual examples  using a python package called  \textit{gender-bender}.
    \item Removing the unused columns in the dataset.
    \item Subsampling the Jigsaw dataset from $1.8$M examples to $125$K examples.
\end{itemize}
The following code shows the pre-processing applied to the $3$ datasets:
\begin{lstlisting}[language=Python, frame = single]
import pandas as pd
from sklearn.model_selection import train_test_split
import numpy as np
from gender_bender import gender_bend


# Pre-processing for the Twitter dataset
df = pd.read_csv(open("./data/waseem.csv"))
df = pd.DataFrame(df).fillna("")
df['Class'] = df['class'].apply(lambda x: 1 if x=="sexism" else 0)
df=df.drop(columns="Unnamed: 0")
df=df.drop(columns="index")
df=df.drop(columns="norm")
df=df.drop(columns="class")
df=df.drop(columns="Unnamed: 0.1")

df_train, df_valid_and_test = train_test_split(df, test_size=0.2)
df_valid, df_test = train_test_split(df_valid_and_test, test_size=0.5)
df_train_gender_swap = df_train.copy()

batch_size = 100

for i in range(int(np.ceil(len(df_train_gender_swap) / batch_size))):
  df_train_gender_swap[df_train_gender_swap.columns[0]][ i * batch_size : (i + 1) * batch_size] = df_train_gender_swap[df_train_gender_swap.columns[0]][ i * batch_size : (i + 1) * batch_size].apply(lambda x: gender_bend(x))


df_train.to_csv('./data/Twitter_train_original_gender.csv')
df_valid.to_csv('./data/Twitter_valid_original_gender.csv')
df_train_gender_swap.to_csv('./data/Twitter_train_gender_swap.csv')    
df_test.to_csv('./data/Twitter_test_original_gender.csv')


# Pre-processing for the Wikipedia dataset
dataset  = "wiki_toxicity"
splits = ["train","valid","test"]
gender = "original_gender"

for split in splits:
    df = pd.read_csv("./data/" + dataset + "_dataset_"+ split+ "_" + gender+".csv")
    df = pd.DataFrame(df).fillna("")
    df['Class'] = 0
    df.Class[df.toxicity>0.5]=1
    
    df=df.drop(columns="rev_id")
    df=df.drop(columns="is_toxic")
    df=df.drop(columns="logged_in")
    df=df.drop(columns="sample")
    df=df.drop(columns="split")
    df=df.drop(columns="year")
    df=df.drop(columns="ns")
    df=df.drop(columns="toxicity")

    df.to_csv("./data/" + dataset + "_"+ split+ "_" + "original_gender" +".csv")
    
    if split == "train":
      df_gender_swap = df.copy()

      for i in range(int(np.ceil(len(df_gender_swap) / batch_size))):
        df_gender_swap[df_gender_swap.columns[0]][ i * batch_size : (i + 1) * batch_size] = df_gender_swap[df_gender_swap.columns[0]][ i * batch_size : (i + 1) * batch_size].apply(lambda x: gender_bend(x))

      df_gender_swap.to_csv("./data/" + dataset + "_"+ split+ "_" + "gender_swap" +".csv")


# Pre-processing for the Jigsaw dataset
df = pd.read_csv(open("./data/train.csv")).iloc[0:125000]
df = pd.DataFrame(df).fillna("")

for column in df.columns:
  if(column!="comment_text" and column!="target"):
      df = df.drop(columns=column)

df['Class'] = df['target'].apply(lambda x: 1 if x>0.5 else 0)

df = df.drop(columns="target")

df_train, df_valid_and_test = train_test_split(df, test_size=0.2)
df_valid, df_test = train_test_split(df_valid_and_test, test_size=0.5)
df_train_gender_swap = df_train.copy()


for i in range(int(np.ceil(len(df_train_gender_swap) / batch_size))):
  df_train_gender_swap[df_train_gender_swap.columns[0]][ i * batch_size : (i + 1) * batch_size] = df_train_gender_swap[df_train_gender_swap.columns[0]][ i * batch_size : (i + 1) * batch_size].apply(lambda x: gender_bend(x))

df_train.to_csv('./data/Jigsaw_train_original_gender.csv')
df_valid.to_csv('./data/Jigsaw_valid_original_gender.csv')
df_train_gender_swap.to_csv('./data/Jigsaw_train_gender_swap.csv')    
df_test.to_csv('./data/Jigsaw_test_original_gender.csv')
\end{lstlisting}
\subsection{Conducting and analyzing experiments}
In this section, we discuss how to run the code to obtain the results of the experiments in the main paper. Running the experiments requires computing the $EL2N$, \textit{forget score}, $GraNd$, and \textit{GE} scores .

\subsubsection{Computing the scores}\label{sec:compute_scores}
The following command is used to compute the \textit{GE} scores for the Twitter dataset using a BERT model and a seed of $1$ when the state of early training is set to $1$ epoch:
 \begin{lstlisting}[language=bash,numbers=none]
python main.py --dataset Twitter --seed 1 --classifier_model bert-base-cased  --CDA_examples_ranking GE --batch_size_biased_model 64 --num_epochs_importance_score 1 --num_epochs_biased_model 1 --batch_size_debiased_model 64  --num_epochs_debiased_model 1 --analyze_results True --compute_importance_scores True
\end{lstlisting}
To account for different datasets, models, and scores, the same command could be run while changing the arguments as shown in Table \ref{tab:computing_the_score}.

\begin{table*}[h!] 
\centering
\begin{tabular}{llllllll}
\hline
\textbf{Argument} & \textbf{Values}\\
\hline
\centering

Dataset                                       & $\in$ $\{\textrm{Twitter}, \textrm{Wiki}, \textrm{Jigsaw}\}$         \\ 
Model                                         & $\in$ $\{\textrm{bert-base-cased}, \textrm{roberta-base}\}$  \\ 
Data augmentation examples ranking            &  $\in$ \{\textrm{\textit{GE}}, \textrm{EL2N}, \textrm{\textit{forget score}}, \textrm{GraNd} \} \\ 
\hline
\end{tabular}
\caption{This tables shows how to change the arguments for the command provided in \ref{sec:compute_scores} to compute different scores for different models and datasets.
}
\label{tab:computing_the_score}
\end{table*}

\begin{table*}[h!] 
\centering
\begin{tabular}{llllllll}
\hline
\textbf{Argument} & \textbf{Values}\\
\hline
\centering

Dataset                                       & $\in$ $\{\textrm{Twitter}, \textrm{Wiki}, \textrm{Jigsaw}\}$         \\ 
Model                                         & $\in$ $\{\textrm{bert-base-cased}, \textrm{roberta-base}\}$  \\ 
Method            &  $\in$ $\{\textrm{data augmentation}, \textrm{data substitution} \}$ \\ 
Data substitution ratio            &  $0.5$ \\ 
Data augmentation ratio            &  $1$ \\ 
\hline
\end{tabular}
\caption{This table shows how to change the arguments for the command provided in \ref{sec:compute_baselines} to measure the performance and bias of the baselines (CDS and CDA) for different models and datasets.
}
\label{tab:baselines}
\end{table*}

\subsubsection{Measuring the performance and bias of the baselines:}\label{sec:compute_baselines}
The following command is used to do debiasing using data substitution for the Wikipedia dataset with a BERT model and a seed of $4$:
 \begin{lstlisting}[language=bash,numbers=none]
python main.py --dataset Wiki --seed 4 --classifier_model bert-base-cased  --method data_substitution --data_substitution_ratio 0.5 --batch_size_biased_model 32  --num_epochs_biased_model 15 --batch_size_debiased_model 32  --num_epochs_debiased_model 15 --analyze_results True 
\end{lstlisting}
To account for different datasets, models, and methods (CDS and CDA), the same command could be run while changing these arguments as shown in Table \ref{tab:baselines}.

\subsubsection{Experiment 1: Verifying that \textit{GE} score reflects the importance of counterfactual examples}\label{sec:experiment_1}
The following command is used to do debiasing using data augmentation for the Twitter dataset using a RoBERTa model and a seed of $3$ using $75$$\%$ of the counterfactual examples that are ranked according to the EL2N score:
 \begin{lstlisting}[language=bash,numbers=none]
python main.py --dataset Twitter --classifier_model roberta-base --seed 3 --data_augmentation_ratio 0.75 --CDA_examples_ranking EL2N --batch_size_biased_model 64 --num_epochs_biased_model 15 --batch_size_debiased_model 64 --method data_augmentation --num_epochs_debiased_model 15 --analyze_results True

\end{lstlisting}
To account for different datasets, models, as well as the ratios and rankings of the counterfactual examples, the same command could be run while changing these arguments as shown in Table \ref{tab:experiment_1}.

\begin{table*}[h!] 
\centering
\begin{tabular}{llllllllll}
\hline

\textbf{Argument} & \textbf{Values}\\
\hline
\centering

Dataset                                       & $\in$ $\{\textrm{Twitter}, \textrm{Wiki}, \textrm{Jigsaw}\}$         \\ 
Model                                         & $\in$ $\{\textrm{bert-base-cased}, \textrm{roberta-base}\}$  \\ 
Data augmentation examples ranking            &  $\in$ $\{\textrm{\textit{GE}}, \textrm{EL2N}, \textrm{\textit{forget score}}, \textrm{GraNd}, \textrm{random}\}$ \\ 
Data augmentation ratio                       &  $\in$ \{0,0.1,0.2,0.3,0.4,0.5,0.75,1\} \\

\end{tabular}
\caption{This table shows how to change the arguments for the command provided in \ref{sec:experiment_1} to run experiment 1 for different datasets, models, as well as rankings and ratios of the counterfactual examples.
}
\label{tab:experiment_1}
\end{table*}

\subsubsection{Experiment 2: Finding the best trade-off between fairness and performance}\label{sec:experiment_2}
The following command is used to do debiasing with $30$$\%$ of the factual examples and $20$$\%$ of the counterfactual examples using the proposed healthy random ranking for the Jigsaw dataset with RoBERTa model and a seed of $5$:
 \begin{lstlisting}[language=bash,numbers=none]

python main.py --batch_size_biased_model 32 --seed 5 --num_epochs_biased_model 10 --dataset Jigsaw --batch_size_debiased_model 32 --method data_diet --num_epochs_debiased_model 10 --classifier_model roberta-base --max_length 40 --analyze_results True --data_diet_factual_ratio 0.3 --data_diet_counterfactual_ratio 0.2  --data_diet_examples_ranking healthy_random

\end{lstlisting}
To account for different datasets, models, rankings, and ratios of the factual and counterfactual examples, the same command could be run by changing these arguments as shown in Table \ref{tab:experiment_2}.

\begin{table*}[h] 
\centering
\begin{tabular}{llllllllll}
\hline
\textbf{Argument} & \textbf{Values}\\
\hline
\centering

Dataset                                       & $\in$ $\{\textrm{Twitter}, \textrm{Wiki}, \textrm{Jigsaw}\}$         \\ 
Model                                         & $\in$ $\{\textrm{bert-base-cased}, \textrm{roberta-base}\}$  \\ 
Data diet examples ranking            &  $\in$ $\{\textrm{Healthy random}, \textrm{Unhealthy random}, \textrm{random}, \textrm{Vanilla \textit{GE}}\}$ \\ 
Ratio of factual examples                       &  $\in$ \{0.3,0.4,0.5\} \\ 
Ratio of counterfactual examples                &  $\in$ \{0,0.1,0.2,0.3,0.4,0.5\} \\

\end{tabular}
\caption{This table shows how to change the arguments for the command provided in \ref{sec:experiment_2} to run experiment 2 for different datasets, models, rankings as well as ratios for the factual and counterfactual examples.
}
\label{tab:experiment_2}
\end{table*}

\subsection{Computing infrastructure}

All the experiments were conducted using 1 Tesla P100-PCIE-16GB GPU. The code is written entirely in Python 3 and the operating system used is Linux. All the packages required to run the code are provided in the $requirements.txt$ file in our code.

\end{document}